\documentclass{ws-nmnc}
\usepackage{latexsym}
\begin{document}

\markboth{Haibin Wang;Andr\'{e} Rogatko;Florentin Smarandache; Rajshekhar Sunderraman}{A Neutrosophic Description Logic}

\catchline{}{}{}{}{}
%%%%%%%%%%%%%%%%%%%%%%%%%%%%%%%%%%%%%%%%%%%%%%%%%%%%%%%%%%%%%%%%%%%%

\title{A Neutrosophic Description Logic}
\author{Haibin Wang}
\address{Biostatistics Research and Informatics Core, Winship Cancer Institute\\Atlanta, GA 30322, USA\\
hwang25@emory.edu}  
\author{Andr\'{e} Rogatko}
\address{Biostatistics Research and Informatics Core, Winship Cancer Institute\\Atlanta, GA 30322, USA\\
andre.rogatko@emoryhealthcare.org}
\author{Florentin Smarandache}
\address{Department of Mathematics and Science, University of New Mexico\\Gallup, NM 87301, USA\\
smarand@unm.edu}
 \author{Rajshekhar Sunderraman}
\address{Department of Computer Science, Georgia State University\\Atlanta, GA 30303, USA\\
raj@cs.gsu.edu}

\maketitle

\begin{history}
\received{Day Month Year}
\revised{Day Month Year}
\end{history}

\begin{abstract}
Description Logics (DLs) are appropriate, widely used, logics for managing 
structured knowledge. They allow reasoning about individuals and 
concepts, {\em i.e.} set of individuals with common properties. Typically, DLs
are limited to dealing with crisp, well defined concepts. That is, concepts
for which the problem whether an individual is an instance of it is a yes/no
question. More often than not, the concepts encountered in the real world do
not have a precisely defined criteria of membership: we may say that an 
individual is an instance of a concept only to a certain degree, depending on
the individual's properties. The DLs that deal with such fuzzy concepts are 
called fuzzy DLs. In order to deal with fuzzy, incomplete, indeterminate
and inconsistent concepts, we need to extend the capabilities of fuzzy DLs
further. 

In this paper we will present an  extension of fuzzy ${\cal ALC}$, combining
Smarandache's neutrosophic logic with a classical DL. In particular, concepts
become neutrosophic (here neutrosophic means fuzzy, incomplete, indeterminate
and inconsistent), thus, reasoning about such 
neutrosophic concepts is supported. We will define its syntax, its semantics,
describe its properties and present a constraint propagation calculus for 
reasoning in it. 

\end{abstract}

\section{Introduction}
The modelling and reasoning with  uncertainty and imprecision is an 
important research topic in the Artificial Intelligence community.
Almost all the real world knowledge is imperfect. 
A lot of works have been carried out to extend
existing knowledge-based systems to deal with such imperfect information, 
resulting in a number of concepts being investigated, a number of
problems being identified and a number of solutions being developed \cite{1,2,3,4}.

{\em Description Logics} (DLs) have been utilized in building a large 
amount of knowledge-based systems. 
DLs are a logical reconstruction
of the so-called frame-based knowledge representation languages, with the aim
of providing a simple well-established Tarski-style declarative semantics to
capture the meaning of the most popular features of structured representation
of knowledge. A main point is that DLs are considered as to be attractive
logics in knowledge based applications as they are a good compromise between
expressive power and computational complexity.

Nowadays, a whole family of knowledge representation systems has been build
using DLs, which differ with respect to their expressiveness, their complexity
and the completeness of their algorithms, and they have been used for building
a variety of applications \cite{5,6,7,8}.

The classical DLs can only deal with crisp, well defined concepts. That is,
concepts for which the problem whether an individual is an instance of it is
a yes/no question. More often than not, the concepts encountered in the real
world do not have a precisely defined criteria of membership. There are many
works attempted to extend the DLs using fuzzy set 
theory \cite{9,10,11,12,13,14}. These fuzzy
DLs can only deal with fuzzy concepts but not incomplete, indeterminate, and
inconsistent concepts (neutrosophic concepts). For example, "Good Person" is a
neutrosophic concepts, in the sense that by different subjective opinions,
the truth-membership degree of tom is good person is 0.6, and the falsity-membership degree of tom is good person is 0.6, which is inconsistent, or the truth-
membership degree of tom is good person is 0.6, and the falsity-membership
degree of tom is good person is 0.3, which is incomplete.

The set and logic that can model and reason with fuzzy, incomplete, indeterminate, and inconsistent information are called neutrosophic set and neutrosophic
logic, respectively \cite{15,16}. In Smarandache's neutrosophic set 
theory,a 
neutrosophic set $A$ defined on universe of discourse $X$, associates 
each element $x$ in $X$ with three membership functions: truth-membership
function $T_A(x)$, indeterminacy-membership function $I_A(x)$, and 
falsity-membership function $F_A(x)$, where $T_A(x), I_A(x), F_A(x)$ are real
standard or non-standard subsets of $]^-0,1^+[$, and $T_A(x), I_A(x), F_A(x)$
are independent. For simplicity, in this paper, we will extend Straccia's fuzzy 
DLs \cite{9,11} with 
neutrosophic logic, called neutrosophic DLs, where we only use two components 
$T_A(x)$ and $F_A(x)$, with $T_A(x) \in [0,1], F_A(x) \in [0,1], 0 \leq T_A(x)
+ F_A(x) \leq 2$. The neutrosophic DLs is based on
the DL ${\cal ALC}$, a significant and expressive representative of the various
DLs. This allows us to adapt it easily to the different DLs presented in the 
literature. Another important point is that we will show that the additional
expressive power has no impact from a computational complexity point of view.
The neutrosophic ${\cal ALC}$ is a strict generalization of fuzzy ${\cal ALC}$,
in the sense that every fuzzy concept and fuzzy terminological axiom can be 
represented by a 
corresponding neutrosophic concept and neutrosophic terminological axiom, 
but not {\em vice versa}.

The rest of paper is organized as follows. In the following section we first
introduce Straccia's ${\cal ALC}$. In section 3 we extend ${\cal ALC}$ to the neutrosophic case and discuss some properties in Section 4, while in Section 5 we will
present a constraint propagation calculus for reasoning in it. Section 6 
concludes and proposes future work. 
 
\section{A Quick Look to Fuzzy ${\cal ALC}$}
We assume three alphabets of symbols, called {\em atomic concepts} 
(denoted by $A$), {\em atomic roles} (denoted by $R$) and {\em individuals}
(denoted by $a$ and $b$). \footnote{Through this work we assume that every
metavariable has an optional subscript or superscript.}

A {\em concept} (denoted by $C$ or $D$) of the language ${\cal ALC}$ is built
out of atomic concepts according to the following syntax rules:

\begin{table}[htb]
\begin{center}
\begin{tabular}{llrl}
$C, D$ & $\longrightarrow$ & $\top |$ & (top concept) \\
       &               & $\bot |$ & (bottom concept) \\
       &               & $A |$ & (atomic concept) \\
       &               & $C \sqcap D |$ & (concept conjunction) \\
       &               & $C \sqcup D |$ & (concept disjunction) \\
       &               & $\neg C |$ & (concept negation) \\
       &               & $\forall R.C |$ & (universal quantification) \\
       &               & $\exists R.C$ & (existential quantification)
\end{tabular}
\end{center}
\end{table}

Fuzzy DL extends classical DL under the framework of Zadeh's fuzzy sets \cite{17}.A {\em fuzzy set} $S$ with respect to an universe $U$ is characterized by a
{\em membership function} $\mu_S : U \rightarrow [0, 1]$, assigning an $S$-membership degree, $\mu_S(u)$, to each element $u$ in $U$.
In fuzzy DL, ({\em i}) a concept $C$, rather than being interpreted as a 
classical set, will be interpreted as a fuzzy set and, thus, concepts become
{\em fuzzy}; and, consequently, ({\em ii}) the statement ``$a$ is $C$", 
{\em i.e.} $C(a)$, will have a truth-value in $[0, 1]$ given by the degree
of membership of being the individual $a$ a member of the fuzzy set $C$. 

\subsection{Fuzzy Interpretation}
A {\em fuzzy interpretation} is now a pair ${\cal I} = (\Delta^{\cal I}, .^{\cal I})$, where $\Delta^{\cal I}$ is, as for the crisp case, the {\em domain},
whereas $.^{\cal I}$ is an {\em interpretation function} mapping
\begin{enumerate}
\item individual as for the crisp case, {\em i.e.} $a^{\cal I} \neq b^{\cal I}$, if $a \neq b$;
\item a concept $C$ into a membership function $C^{\cal I}:\Delta^{\cal I} \rightarrow [0, 1]$;
\item a role $R$ into a membership function $R^{\cal I}:\Delta^{\cal I} \times \Delta^{\cal I} \rightarrow [0, 1]$.
\end{enumerate}
If $C$ is a concept then $C^{\cal I}$ will naturally be interpreted as the
{\em membership degree function} of the fuzzy concept (set) $C$ w.r.t. 
${\cal I}$, {\em i.e.} if $d \in \Delta^{\cal I}$ is an object of the domain
$\Delta^{\cal I}$ then $C^{\cal I}(d)$ gives us the degree of being the object
$d$ an element of the fuzzy concept $C$ under the interpretation ${\cal I}$.
Similarly for roles. Additionally, the interpretation function $.^{\cal I}$ has
to satisfy the following equations: for all $d \in \Delta^{\cal I}$,
\begin{table}[htb]
\begin{center}
\begin{tabular}{lcl}
$\top^{\cal I}(d)$ &=& $1$ \\
$\bot^{\cal I}(d)$ &=& $0$ \\
$(C \sqcap D)^{\cal I}(d)$ &=& $\min\{C^{\cal I}(d), D^{\cal I}(d)\}$ \\
$(C \sqcup D)^{\cal I}(d)$ &=& $\max\{C^{\cal I}(d), D^{\cal I}(d)\}$ \\
$(\neg C)^{\cal I}(d)$      &=& $1 - C^{\cal I}(d)$ \\
$(\forall R.C)^{\cal I}(d)$ &=& $\inf_{d' \in \Delta^{\cal I}}\{\max\{1 - R^{\cal I}(d, d'), C^{\cal I}(d')\}\}$ \\
$(\exists R.C)^{\cal I}(d)$ &=& $\sup_{d' \in \Delta^{\cal I}}\{\min\{R^{\cal I}(d, d'), C^{\cal I}(d')\}\}$. \\
\end{tabular}
\end{center}
\end{table}

We will say that two concepts $C$ and $D$ are said to be {\em equivalent} 
(denoted by $C \cong D$) when $C^{\cal I} = D^{\cal I}$ for all interpretation
${\cal I}$. As for the crisp non fuzzy case, dual relationships between 
concepts hold: {\em e.g.} $\top \cong \neg \bot$, $(C \sqcap D) \cong \neg (\neg C \sqcup \neg D)$ and $(\forall R.C) \cong \neg (\exists R. \neg C)$.

\subsection{Fuzzy Assertion}
A {\em fuzzy assertion} (denoted by $\psi$) is an expression having one of the
following forms $\langle \alpha \geq n \rangle$ or $\langle \alpha \leq m \rangle$, where $\alpha$ is an ${\cal ALC}$ assertion, $n \in (0, 1]$ and $m \in [0, 1)$. 
From a semantics point of view, a fuzzy assertion 
$\langle \alpha \leq n \rangle$ constrains the truth-value of $\alpha$ to be 
less or equal to $n$ (similarly for $\geq$). Consequently, {\em e.g.} 
$\langle$ ({\bf Video} $\sqcap \mbox{ } \exists${\bf About.Basket})({\bf v1}) 
$\geq 0.8 \rangle$ 
states that video {\bf v1} is likely about basket. Formally, an interpretation
${\cal I}$ {\em satisfies} a fuzzy assertion $\langle C(a) \geq n \rangle$ 
(resp. $\langle R(a, b) \geq n \rangle$) iff $C^{\cal I}(a^{\cal I}) \geq n$
(resp. $R^{\cal I}(a^{\cal I}, b^{\cal I}) \geq n$). Similarly, an 
interpretation ${\cal I}$ {\em satisfies} a fuzzy assertion 
$\langle C(a) \leq n \rangle$ (resp. $\langle R(a, b) \leq n \rangle$) iff 
$C^{\cal I}(a^{\cal I}) \leq n$ (resp. $R^{\cal I}(a^{\cal I}, b^{\cal I}) \leq n$). Two fuzzy assertion $\psi_1$ and $\psi_2$ are said to be {\em equivalent}
(denoted by $\psi_1 \cong \psi_2$) iff they are satisfied by the same set of 
interpretations. An {\em atomic fuzzy assertion} is a fuzzy assertion involving
an atomic assertion (assertion of the form $A(a)$ or $R(a,b)$).

\subsection{Fuzzy Terminological Axiom}
From a syntax point of view, a {\em fuzzy terminological axiom} (denoted by
$\tilde{\tau}$ is either a fuzzy concept specialization or a fuzzy concept
definition. A {\em fuzzy concept specialization} is an expression of the form
$A \prec C$, where $A$ is an atomic concept and $C$ is a concept. On the 
other hand, a {\em fuzzy concept definition} is an expression of the form
$A :\approx C$, where $A$ is an atomic concept and $C$ is a concept. From
a semantics point of view, a fuzzy interpretation ${\cal I}$ {\em satisfies}
a fuzzy concept specialization $A \prec C$ iff
\begin{equation}
\forall d \in \Delta^{\cal I}, A^{\cal I}(d) \leq C^{\cal I}(d),
\end{equation}
whereas ${\cal I}$ {\em satisfies} a fuzzy concept definition $A :\approx C$ iff
\begin{equation}
\forall d \in \Delta^{\cal I}, A^{\cal I}(d) = C^{\cal I}(d).
\end{equation}

\subsection{Fuzzy Knowlege Base, Fuzzy Entailment and Fuzzy Subsumption}
A {\em fuzzy knowledge base} is a finite set of fuzzy assertions and fuzzy
terminological axioms. $\Sigma_A$ denotes the set of fuzzy assertions in 
$\Sigma$, $\Sigma_T$ denotes the set of fuzzy terminological axioms in $\Sigma$
(the terminology), if $\Sigma_T = \emptyset$ then $\Sigma$ is {\em purely assertional}, and we will assume that a terminology $\Sigma_T$ is such that no concept
$A$ appears more than once on the left hand side of a fuzzy terminological
axiom $\tilde{\tau} \in \Sigma_T$ and that no cyclic definitions are present
in $\Sigma_T$. 

An interpretation ${\cal I}$ {\em satisfies} (is {\em a model of}) a set of 
fuzzy $\Sigma$ iff ${\cal I}$ satisfies each element of $\Sigma$. A fuzzy
KB $\Sigma$ {\em fuzzy entails} a fuzzy assertion $\psi$ (denoted by $\Sigma
\models_f \psi$) iff every model of $\Sigma$ also satisfies $\psi$.

Furthermore, let $\Sigma_T$ be a terminology and let $C, D$ be two concepts.
We will say that $D$ {\em fuzzy subsumes} $C$ w.r.t. $\Sigma_T$ (denoted by
$C \preceq_{\Sigma_T} D$) iff for every model ${\cal I}$ of $\Sigma_T$,
$\forall d \in \Delta^{\cal I}, C^{\cal I}(d) \leq D^{\cal I}(d)$ holds. 

\section{A Neutrosophic DL}
Our neutrosophic extension directly relates to Smarandache's work on 
neutrosophic sets \cite{15,16}. 
A neutrosophic set $S$ defined on universe of discourse $U$, associates
each element $u$ in $U$ with three membership functions: truth-membership
function $T_S(u)$, indeterminacy-membership function $I_S(u)$, and
falsity-membership function $F_S(u)$, where $T_S(u), I_S(u), F_S(u)$ are real
standard or non-standard subsets of $]^-0,1^+[$, and $T_S(u), I_S(u), F_S(u)$
are independent. For simplicity, here we only use two components 
$T_S(u)$ and $F_S(u)$, with $T_S(u) \in [0,1], F_S(u) \in [0,1], 0 \leq T_S(u)
+ F_S(u) \leq 2$. It is easy to extend our method to include 
indeterminacy-membership function. 
$T_S(u)$ gives us an estimation of degree of $u$ belonging to $U$ and
$F_S(u)$ gives us an estimation of degree of $u$ not belonging to $U$.
$T_S(u) + F_S(u)$ can be 1 (just as in classical fuzzy sets theory).
But it is not necessary. If $T_S(u) + F_S(u) < 1$, for all $u$ in $U$, 
we say the set $S$ is 
incomplete, if
$T_S(u) + F_S(u) > 1$, for all $u$ in $U$, we say the set $S$ is inconsistent. 
According
to Wang \cite{16}, the truth-membership function and falsity-membership function
has to satisfy three restrictions: for all $u \in U$ and for all neutrosophic
sets $S_1, S_2$ with respect to $U$
\[
T_{S_1 \cap S_2}(u) = \min\{T_{S_1}(u), T_{S_2}(u)\}, F_{S_1 \cap S_2}(u) = \max\{F_{S_1}(u), F_{S_2}(u)\}
\]
\[
T_{S_1 \cup S_2}(u) = \max\{T_{S_1}(u), T_{S_2}(u)\}, F_{S_1 \cup S_2}(u) = \min\{F_{S_1}(u), F_{S_2}(u)\}
\]
\[
T_{\overline{S_1}}(u) = F_{S_1}(u), F_{\overline{S_1}}(u) = T_{S_1}(u),
\]
where $\overline{S_1}$ is the complement of $S_1$ in $U$.
Wang \cite{16} gives the definition of $N$-norm and $N$-conorm of neutrosophic 
sets, $\min$ and $\max$ is only one of the choices. In general case, they may
be the simplest and the best.  

When we switch to neutrosophic logic, the notion of degree of truth-membership
$T_S(u)$ of an element $u \in U$ w.r.t. the neutrosophic set $S$ over $U$ is
regarded as the {\em truth-value} of the statement ``$u$ is $S$", and
the notion of degree of falsity-membership $F_S(u)$ of an element $u \in U$
w.r.t. the neutrosophic set $S$ over $U$ is regarded as the {\em falsity-value}
of the statement ``$u$ is $S$". Accordingly, in our neutrosophic DL,
({\em i}) a concept $C$, rather than being interpreted as a fuzzy set, will
be interpreted as a neutrosophic set and, thus, concepts become {\em imprecise}
(fuzzy, incomplete, and inconsistent); and, consequently, ({\em ii}) the 
statement ``$a$ is $C$", {\em i.e.} $C(a)$ will have a truth-value in $[0, 1]$
given by the degree of truth-membership of being the individual $a$ a member
of the neutrosophic set $C$ and a falsity-value in $[0, 1]$ given by the 
degree of falsity-membership of being the individual $a$ not a member of the
neutrosophic set $C$.

\subsection{Neutrosophic Interpretation}
A {em neutrosophic interpretation} is now a tuple ${\cal I} = (\Delta^{\cal I},
(\cdot)^{\cal I}, |\cdot|^t, |\cdot|^f)$, where $\Delta^{\cal I}$ is, as for
the fuzzy case, the {\em domain}, and
\begin{enumerate}
\item $(\cdot)^{\cal I}$ is an {\em interpretation function} mapping
  \begin{enumerate}
      \item individuals as for the fuzzy case, {\em i.e.} $a^{\cal I} \neq b^{\cal I}$, if $a \neq b$;
      \item a concept $C$ into a membership function $C^{\cal I} : \Delta^{\cal I} \rightarrow [0, 1] \times [0, 1]$;
      \item a role $R$ into a membership function $R^{\cal I} : \Delta^{\cal I} \times \Delta^{\cal I} \rightarrow [0, 1] \times [0, 1]$.
   \end{enumerate}
\item $|\cdot|^t$ and $|\cdot|^f$ are {\em neutrosophic valuation}, {\em i.e.}
$|\cdot|^t$ and $|\cdot|^f$ map
   \begin{enumerate}
      \item every atomic concept into a function from $\Delta^{\cal I}$ to $[0, 1]$;
      \item every atomic role into a function from $\Delta^{\cal I} \times \Delta^{\cal I}$ to $[0, 1]$.
   \end{enumerate}
\end{enumerate}

If $C$ is a concept then $C^{\cal I}$ will naturally be interpreted as a pair
of membership functions $\langle |C|^t, |C|^f \rangle$ of the neutrosophic
concept (set) $C$ w.r.t. ${\cal I}$, {\em i.e.} if $d \in \Delta^{\cal I}$ is
an object of the domain $\Delta^{\cal I}$ then $C^{\cal I}(d)$ gives us the
degree of being the object $d$ an element of the neutrosophic concept $C$ and
the degree of being the object $d$ not an element of the neutrosophic concept
$C$ under the interpretation ${\cal I}$. Similarly for roles. Additionally,
the interpretation function $(\cdot)^{\cal I}$ has to satisfy the following
equations: for all $d \in \Delta^{\cal I}$,

\begin{table}[htb]
\begin{center}
\begin{tabular}{lcl}
$\top^{\cal I}(d)$ &=& $\langle 1, 0 \rangle$ \\
$\bot^{\cal I}(d)$ &=& $\langle 0, 1 \rangle$ \\
$(C \sqcap D)^{\cal I}(d)$ &=& $\langle \min\{|C|^t(d), |D|^t(d)\}, \max\{|C|^f(d), |D|^f(d)\} \rangle$ \\
$(C \sqcup D)^{\cal I}(d)$ &=& $\langle \max\{|C|^t(d), |D|^t(d)\}, \min\{|C|^f(d), |D|^f(d)\} \rangle$ \\
$(\neg C)^{\cal I}(d)$ &=& $\langle |C|^f(d), |C|^t \rangle$ \\
$(\forall R.C)^{\cal I}(d)$ &=& $\langle \inf_{d' \in \Delta^{\cal I}}\{\max\{|R|^f(d, d'), |C|^t(d')\}\}, \sup_{d' \in \Delta^{\cal I}}\{\min\{|R|^t(d, d'), |C|^f(d')\}\} \rangle$ \\
$(\exists R.C)^{\cal I}(d)$ &=& $\langle \sup_{d' \in \Delta^{\cal I}}\{\min\{|R|^t(d, d'), |C|^t(d')\}\}, \inf_{d' \in \Delta^{\cal I}}\{\max\{|R|^f(d, d'), |C|^f(d')\}\} \rangle$
\end{tabular}
\end{center}
\end{table} 
Note that the semantics of $\forall R.C$
\begin{equation}
(\forall R.C)^{\cal I}(d) = \langle \inf_{d' \in \Delta^{\cal I}}\{\max\{|R|^f(d, d'), |C|^t(d')\}\}, \sup_{d' \in \Delta^{\cal I}}\{\min\{|R|^t(d, d'), |C|^f(d')\}\} \rangle
\end{equation}
is the result of viewing $\forall R.C$ as the open first order formula
$\forall y. \neg F_R(x, y) \vee F_C(y)$, where the universal quantifier 
$\forall$ is viewed as a conjunction over the elements of the domain.
Similarly, the semantics of $\exists R.C$
\begin{equation}
(\exists R.C)^{\cal I}(d) = \langle \sup_{d' \in \Delta^{\cal I}}\{\min\{|R|^t(d, d'), |C|^t(d')\}\}, \inf_{d' \in \Delta^{\cal I}}\{\max\{|R|^f(d, d'), |C|^f(d')\}\} \rangle
\end{equation}
is the result of viewing $\exists R.C$ as the open first order formula
$\exists y. F_R(x, y) \wedge F_C(y)$ and the existential quantifier $\exists$
is viewed as a disjunction over the elements of the domain. Moreover, 
$|\cdot|^t$ and $|\cdot|^f$ are extended to complex concepts as follows: for 
all $d \in \Delta^{\cal I}$
\begin{table}[htb]
\begin{center}
\begin{tabular}{lcl}
$|C \sqcap D|^t(d)$ &=& $\min\{|C|^t(d), |D|^t(d)\}$ \\
$|C \sqcap D|^f(d)$ &=& $\max\{|C|^f(d), |D|^f(d)\}$ \\ \\
$|C \sqcup D|^t(d)$ &=& $\max\{|C|^t(d), |D|^t(d)\}$ \\
$|C \sqcup D|^f(d)$ &=& $\min\{|C|^f(d), |D|^f(d)\}$ \\ \\
$|\neg C|^t(d)$     &=& $|C|^f(d)$ \\
$|\neg C|^f(d)$     &=& $|C|^t(d)$ \\ \\
$|\forall R.C|^t(d)$ &=& $\inf_{d' \in \Delta^{\cal I}}\{\max\{|R(d, d')|^f, |C|^t(d)\}\}$ \\
$|\forall R.C|^f(d)$ &=& $\sup_{d' \in \Delta^{\cal I}}\{\min\{|R(d, d')|^t, |C|^f(d)\}\}$ \\ \\
$|\exists R.C|^t(d)$ &=& $\sup_{d' \in \Delta^{\cal I}}\{\min\{|R(d, d')|^t, |C|^t(d)\}\}$ \\
$|\exists R.C|^f(d)$ &=& $\inf_{d' \in \Delta^{\cal I}}\{\max\{|R(d, d')|^f, |C|^f(d)\}\}$
\end{tabular}
\end{center}
\end{table}

We will say that two concepts $C$ and $D$ are said to be {\em equivalent}
(denoted by $C \cong^n D$) when $C^{\cal I} = D^{\cal I}$ for all 
interpretation ${\cal I}$. As for the fuzzy case, dual relationships between
concepts hold: {\em e.g.} $\top \cong^n \neg \bot, (C \sqcap D) \cong^n \neg(\neg C \sqcup \neg D)$ and $(\forall R.C) \cong^n \neg (\exists R.\neg C)$.
%\bibliography{ref}

\subsection{Neutrosophic Assertion}
A {\em neutrosophic assertion} (denoted by $\varphi$) is an expression having
one of the following form $\langle \alpha:\geq n, \leq m \rangle$ or
$\langle \alpha:\leq n, \geq m \rangle$, where $\alpha$ is an ${\cal ALC}$
assertion, $n \in [0, 1]$ and $m \in [0, 1]$. From a semantics point of view,
a neutrosophic assertion $\langle \alpha:\geq n, \leq m \rangle$ constrains
the truth-value of $\alpha$ to be greater or equal to $n$ and falsity-value
of $\alpha$ to be less or equal to $m$ (similarly for $\langle \alpha:\leq n, \geq m \rangle$). Consequently, {\em e.g.} $\langle$({\bf Poll} $\sqcap \mbox{ }\exists${\bf Support.War\_x})({\bf p1}) $:\geq 0.8, \leq 0.1 \rangle$ states that
poll {\bf p1} is close to support War\_x. Formally, an interpretation 
${\cal I}$
{\em satisfies} a neutrosophic assertion 
$\langle \alpha:\geq n, \leq m \rangle$ 
(resp. $\langle R(a, b):\geq n, \leq m \rangle$) iff $|C|^t(a^{\cal I}) \geq n$
and $|C|^f(a^{\cal I}) \leq m$ (resp. $|R|^t(a^{\cal I}, b^{\cal I}) \geq n$ 
and $|R|^f(a^{\cal I}, b^{\cal I}) \leq m$). Similarly, an interpretation
${\cal I}$ {\em satisfies} a neutrosophic assertion 
$\langle \alpha:\leq n, \geq m \rangle$ 
(resp. $\langle R(a, b):\leq n, \geq m \rangle$) iff $|C|^t(a^{\cal I}) \leq n$
and $|C|^f(a^{\cal I}) \geq m$ (resp. $|R|^t(a^{\cal I}, b^{\cal I}) \leq n$
and $|R|^f(a^{\cal I}, b^{\cal I}) \geq m$). Two fuzzy assertion $\varphi_1$
and $\varphi_2$ are said to be {\em equivalent} (denoted by 
$\varphi_1 \cong^n \varphi_2$) iff they are satisfied by the same set of 
interpretations. Notice that $\langle \neg C(a): \geq n, \leq m \rangle \cong^n \langle C(a): \leq m, \geq n \rangle$ and $\langle \neg C(a): \leq n, \geq m \rangle \cong^n \langle C(a): \geq m, \leq n \rangle$. 
An {\em atomic neutrosophic
assertion} is a neutrosophic assertion involving an atomic assertion.

\subsection{Neutrosophic Terminological Axiom} 
Neutrosophic terminological axioms we will consider are a natural extension
of fuzzy terminological axioms to the neutrosophic case. From a syntax point of
view, a {\em neutrosophic terminological axiom} (denoted by $\hat{\tau}$) is
either a neutrosophic concept specialization or a neutrosophic concept 
definition. A {\em neutrosophic concept specialization} is an expression of
the form $A \prec^n C$, where $A$ is an atomic concept and $C$ is a concept.
On the other hand, a {\em neutrosophic concept definition} is an expression
of the form $A :\approx^n C$, where $A$ is an atomic concept and $C$ is a 
concept. From a semantics point of view, we consider the natural extension
of fuzzy set to the neutrosophic case \cite{15,16}. A neutrosophic interpretation
${\cal I}$ {\em satisfies} a neutrosophic concept specialization $A \prec^n C$
iff
\begin{equation}
\forall d \in \Delta^{\cal I}, |A|^t(d) \leq |C|^t(d), |A|^f(d) \geq |C|^f(d),
\end{equation} 
whereas ${\cal I}$ {\em satisfies} a neutrosophic concept definition
$A :\approx^n C$ iff
\begin{equation}
\forall d \in \Delta^{\cal I}, |A|^t(d) = |C|^t(d), |A|^f(d) = |C|^f(d).
\end{equation}

\subsection{Neutrosophic Knowledge Base, Neutrosophic Entailment and Neutrosophic Subsumption}
A {\em neutrosophic knowledge base} is a finite set of neutrosophic assertions
and neutrosophic terminological axioms. As for the fuzzy case, with $\Sigma_A$
we will denote the set of neutrosophic assertions in $\Sigma$, with $\Sigma_T$
we will denote the set of neutrosophic terminological axioms in $\Sigma$
(the terminology), if $\Sigma_T = \emptyset$ then $\Sigma$ is {\em purely
assertional}, and we will assume that a terminology $\Sigma_T$ is such that no
concept $A$ appears more than once on the left hand side of a neutrosophic
terminological axiom $\hat{\tau} \in \Sigma_T$ and that no cyclic definitions
are present in $\Sigma_T$.

An interpretation ${\cal I}$ {\em satisfies} (is {\em a model of}) a 
neutrosophic $\Sigma$ iff ${\cal I}$ satisfies each element of $\Sigma$.
A neutrosophic KB $\Sigma$ {\em neutrosophically entails} a neutrosophic assertion
$\varphi$ (denoted by $\Sigma \models^n \varphi$) iff every model of $\Sigma$
also satisfies $\varphi$.
 
Furthermore, let $\Sigma_T$ be a terminology and let $C, D$ be two concepts.
We will say that $D$ {\em neutrosophically subsumes} $C$ w.r.t. $\Sigma_T$
(denoted by $C \preceq_{\Sigma_T}^n D$) iff for every model ${\cal I}$ of
$\Sigma_T$, $\forall d \in \Delta^{\cal I}, |C|^t(d) \leq |D|^t(d)$ and
$|C|^f(d) \geq |D|^f(d)$ holds.

Finally, given a neutrosophic KB $\Sigma$ and an assertion $\alpha$, we define
the {\em greatest lower bound} of $\alpha$ w.r.t. $\Sigma$ (denoted by $glb(\Sigma, \alpha)$) to be $\langle \sup\{n : \Sigma \models^n \langle \alpha:\geq n, \leq m \rangle\}, \inf\{m : \Sigma \models^n \langle \alpha:\geq n, \leq m \rangle\} \rangle$. Similarly, we define the {\em least upper bound} of $\alpha$
with respect to $\Sigma$ (denoted by $lub(\Sigma, \alpha)$) to be $\langle \inf\{n : \Sigma \models^n \langle \alpha:\leq n, \geq m \rangle\}, \sup\{m : \Sigma \models^n \langle \alpha:\leq n, \geq m \rangle\} \rangle$ ($\sup \emptyset = 0, \inf \emptyset = 1$). Determing the $lub$ and the $glb$ is called the 
{\em Best Truth-Value Bound} (BTVB) problem. 

\section{Some Properties}
In this section, we discuss some properties of our neutrosophic ${\cal ALC}$.

\subsection{Concept Equivalence}
The first ones are straightforward: $\neg \top \approx^n \bot, C \sqcap \top \approx^n C, C \sqcup \top \approx^n \top, C \sqcap \bot \approx^n \bot, C \sqcup \bot \approx^n C, \neg\neg C \approx^n C, \neg(C \sqcap D) \approx^n \neg C \sqcup \neg D, \neg(C \sqcup D) \approx^n \neg C \sqcap \neg D, C_1 \sqcap (C_2 \sqcup C_3) \approx^n (C_1 \sqcap C_2) \sqcup (C_1 \sqcap C_3)$ and $C_1 \sqcup (C_2 \sqcap C_3) \approx^n (C_1 \sqcup C_2) \sqcap (C_1 \sqcup C_3)$. 
For concepts involving roles, we have $\forall R.C \approx^n \neg \exists R.\neg C, \forall R.\top \approx^n \top, \exists R.\bot \approx^n \bot$ and
$(\forall R.C) \sqcap (\forall R.D) \approx^n \forall R.(C \sqcap D)$. Please
note that we do not have $C \sqcap \neq C \approx^n \bot$, nor we have
$C \sqcup \neg C \approx^n \top$ and, thus, $(\exists R.C) \sqcap (\forall R.\neg C) \approx^n \bot$ and $(\exists R.C) \sqcup (\forall R.\neg C) \approx^n \top$ do not hold.

\subsection{Entailment Relation}
Of course, $\Sigma \models^n \langle \alpha:\geq n, \leq m \rangle$ iff
$glb(\Sigma, \alpha) = \langle f, g \rangle$ with $f \geq n$ and
$g \leq m$, and similarly $\Sigma \models^n \langle \alpha:\leq n, \geq m \rangle$ iff $lub(\Sigma, \alpha) = \langle f, g \rangle$ with $f \leq n$ and
$g \geq m$. Concerning roles, note that $\Sigma \models^n \langle R(a,b):\geq n, \leq m \rangle$ iff $\langle R(a,b):\geq f, \leq g \rangle \in \Sigma$ with
$f \geq n$ and $g \leq m$. Therefore,
\begin{eqnarray}
glb(\Sigma, R(a, b)) &=& \langle \max\{n : \langle R(a, b):\geq n, \leq m \rangle\ \in \Sigma\}, \nonumber \\
 & & \min\{m : \langle R(a, b):\geq n, \leq m \rangle\ \in \Sigma\} \rangle
\end{eqnarray} 
while the same is not true for the $\langle R(a, b):\leq n, \geq m$ case. While
$\langle R(a,b):\leq f, \geq g \rangle \in \Sigma$ and $f \leq n, g \geq m$
imply $\Sigma \models^n \langle R(a,b):\leq n, \geq m \rangle$, the converse
is false ({\em e.g.} $\{\langle \forall R.A(a):\geq 1, \leq 0 \rangle, \langle A(b):\leq 0, \geq 1 \rangle \} \models^n \langle R(a, b):\leq 0, \geq 1 \rangle$).

Furthermore, from $\Sigma \models^n \langle C(a):\leq n, \geq m \rangle$ iff
$\Sigma \models^n \langle \neg C(a):\geq m, \leq n \rangle$, it follows 
$lub(\Sigma, C(a)) = \langle f, g \rangle$ iff $glb(\Sigma, \neg C(a)) = \langle g, f \rangle$. Therefore, $lub$ can be determined through
$glb$ (and vice versa). The same reduction to $glb$ does not hold for
$lub(\Sigma, R(a, b))$ as $\neg R(a, b)$ is not an expression of our 
language.

{\em Modus ponens on concepts} is supported: if $n > g$ and $m < f$ then
$\{\langle C(a):\geq n, \leq m \rangle, \langle (\neg C \sqcup D)(a):\geq f, \leq g \rangle\} \models^n \rangle D(a):\geq f, \leq g \rangle$ holds.

{\em Modus ponens on roles} is supported: if $n > g$ and $m < f$ then
$\{\langle R(a, b):\geq n, \leq m \rangle, \langle \forall R.D(a):\geq f, \leq g \rangle\} \models^n \langle D(b):\geq f, \leq g \rangle$ and $\{\langle \exists R.C(a):\geq n, \leq m \rangle, \langle \forall R.D(a):\geq f, \leq g \rangle\} \models^n \langle \exists R.(C \sqcap D)(a):\geq \min\{n, f\}, \leq \max\{m, g\} \rangle$ hold. Moreover, $\{\langle \forall R.C(a):\geq n, \leq m \rangle, \langle \forall R.D(a):\geq f, \leq g \rangle\} \models^n \langle \forall (R.(C \sqcap D))(a):\geq \min\{n, f\}, \leq \max\{m, g\} \rangle$ holds.

{\em Modus ponens on specialization} is supported. The following degree bounds
propagation through a taxonomy is supported. If $C \preceq_\Sigma^n D$ then
({\em i}) $\Sigma \cup \{\langle C(a):\geq n, \leq m \rangle\} \models^n \langle D(a):\geq n, \leq m \rangle\}$; and ({\em ii}) $\Sigma \cup \{\langle D(a):\leq n, \geq m \rangle\} \models^n \langle C(a):\leq n, \geq m \rangle$ hold.

\subsection{Soundness and Completeness of the Semantics} 
Our neutrosophic semantics is {\em sound} and {\em complete} w.r.t. 
fuzzy semantics. First we must note that 
the neutrosophic ${\cal ALC}$ is a strict generalization of fuzzy ${\cal ALC}$,
in the sense that every fuzzy concept and fuzzy terminological axiom can be
represented by a
corresponding neutrosophic concept and neutrosophic terminological axiom, 
but not {\em vice versa}. It is easy to verify that,

\begin{proposition}
A classical fuzzy ${\cal ALC}$ can be simulated by a neutrosophic ${\cal ALC}$,
in the way that a fuzzy assertion $\langle \alpha \geq n \rangle$ 
represented by a neutrosophic assertion 
$\langle \alpha:\geq n, \leq 1-n \rangle$,
a fuzzy assertion $\langle \alpha \leq n \rangle$ represented by a
neutrosophic assertion $\langle \alpha:\leq n, \geq 1-n \rangle$ and a fuzzy 
terminological axiom $\tilde{\tau}$ represented by a neutrosophic 
terminological axiom $\hat{\tau}$ in the sense that if ${\cal I}$ is a
fuzzy interpretation then $|C|^t(a) = C^{\cal I}(a)$ and 
$|C|^f(a) = 1-C^{\cal I}(a)$. $\hfill \dashv$ 
\end{proposition}
 
Let us
consider the following transformations $\sharp(\cdot)$ and $\star(\cdot)$ of 
neutrosophic
assertions into fuzzy assertions,
\begin{eqnarray*}
\sharp \langle \alpha:\geq n, \leq m \rangle &\mapsto& \langle \alpha \geq n \rangle, \\
\star \langle \alpha:\geq n, \leq m \rangle &\mapsto& \langle \alpha \leq m \rangle, \\ 
\sharp \langle \alpha:\leq n, \geq m \rangle &\mapsto& \langle \alpha \leq n \rangle, \\
\star \langle \alpha:\leq n, \geq m \rangle &\mapsto& \langle \alpha \geq m \rangle, 
\end{eqnarray*}
We extend $\sharp(\cdot)$ and $\star(\cdot)$ to neutrosophic terminological 
axioms as follows:
$\sharp \hat{\tau} = \tilde{\tau}$ and $\star \hat{\tau} = \tilde{\tau}$. 
Finally, $\sharp \Sigma = \{\sharp \varphi : \varphi \in \Sigma_A\} \cup \{\sharp \hat{\tau} : \hat{\tau} \in \Sigma_T\}$ and $\star \Sigma = \{\star \varphi : \varphi \in \Sigma_A\} \cup \{\star \hat{\tau} : \hat{\tau} \in \Sigma_T\}$.

\begin{proposition}
Let $\Sigma$ be a neutrosophic KB and let $\varphi$ be a neutrosophic assertion
($\langle \alpha:\geq n, \leq m \rangle$ or $\langle \alpha:\leq n, \geq m \rangle$). Then $\Sigma \models^n \varphi$ iff 
$\sharp \Sigma \models \sharp \varphi$ and $\star \Sigma \models \star \varphi$.
$\mbox{  }$ $\hfill \dashv$
\end{proposition}
\begin{proof}
($\Rightarrow$): Let $\varphi$ be $\langle \alpha:\geq n, \leq m \rangle$.
Consider a fuzzy interpretation ${\cal I}$ satisfying $\sharp \Sigma$ and 
${\cal I^{'}}$ satisfying $\star \Sigma$. 
$\langle {\cal I}, {\cal I^{'}} \rangle$ is also a neutrosophic interpretation 
such that $a^{\cal I} = a^{\cal I^{'}}$, $C^{{\cal I}}(a) = |C|^t(a)$ and $C^{{\cal I^{'}}}(a) = |C|^f(a)$, 
$R^{{\cal I}}(d, d') = |R|^t(d, d')$ and 
$R^{{\cal I^{'}}}(d, d') = |R|^f(d, d')$ hold. By induction on the structure 
of a concept $C$ it
can be shown that ${\cal I}$ (${\cal I^{'}}$) satisfies $C(a)$ iff $C^{{\cal I}}(a^{\cal I}) \geq n$ ($C^{{\cal I{'}}}(a^{\cal I^{'}} \geq n$) for fuzzy 
assertion $\langle C(a) \geq n \rangle$ and $C^{\cal I}(a^{\cal I}) \leq n$ 
($C^{\cal I^{'}}(a^{\cal I^{'}}$) for fuzzy assertion 
$\langle C(a) \leq n \rangle$. Similarly
for roles. By the definition of $\sharp(\cdot)$ and $\star(\cdot)$, therefore 
$\langle {\cal I}, {\cal I^{'}} \rangle$ is a
neutrosophic interpretation satisfying $\Sigma$. By hypothesis, 
$\langle {\cal I}, {\cal I^{'}} \rangle$
satisfies $\langle \alpha:\geq n, \leq m \rangle$. Therefore, ${\cal I}$
satisfies $\sharp \varphi$ and ${\cal I^{'}}$ satisfies $\sharp \varphi$. 
The proof is similar for $\varphi = \langle \alpha:\leq n, \geq m \rangle$.

($\Leftarrow$): Let $\varphi$ be $\langle \alpha:\geq n, \leq m \rangle$.
Consider a neutrosophic ${\cal I}$ satisfying $\Sigma$. ${\cal I}$ can be 
regarded as two fuzzy interpretations ${\cal I^{'}}$ and ${\cal I^{"}}$ 
such that $a^{\cal I} = a^{\cal I^{'}} = a^{\cal I^{"}}$, 
$C^{\cal I^{'}}(d) = |C|^t(d)$ and $C^{\cal I^{"}}(d) = |C|^f(d)$, 
$R^{\cal I^{'}}(d, d') = |R|^t(d, d')$ and $R^{\cal I^{"}}(d, d') =|R|^f(d, d')$hold. By induction on the structure of a concept $C$ it can be
shown that ${\cal I}$ satisfies 
$C(a)$ iff $|C|^t(a^{\cal I}) \geq n, |C|^f(a^{\cal I}) \leq m$ for 
neutrosophic assertion $\langle C(a):\geq n, \leq m \rangle$ and 
$|C|^t(a^{\cal I}) \leq n, |C|^f(a^{\cal I}) \geq m$ for
neutrosophic assertion $\langle C(a):\leq n, \geq m \rangle$. Similarly for
roles. By the definition of $\sharp(\cdot)$ and $\star(\cdot)$, therefore, 
${\cal I^{'}}$ is a fuzzy
interpretation satisfying $\sharp \Sigma$ and ${\cal I^{"}}$ satisfying
$\star \Sigma$. By hypothesis, ${\cal I^{'}}$ satisfies
$\sharp \varphi$ and ${\cal I^{"}}$ satisfies $\star \varphi$. And according to the definition of $\sharp(\cdot)$ and $\star(\cdot)$, ${\cal I}$ satisfies $\langle \alpha:\geq n, \leq m \rangle$. The proof is similar for $\varphi = \langle \alpha:\leq n, \geq m \rangle$. $\hfill \Box$ 
\end{proof}

\subsection{Subsumption}
As for the fuzzy case, subsumption between two concepts $C$ and $D$ w.r.t.
a terminology $\Sigma_T$, {\em i.e.} $C \preceq_{\Sigma_T}^n D$, can be
reduced to the case of an empty terminology, {\em i.e.} $C' \preceq_{\emptyset}^n D'$.
\begin{example}
Suppose we have two polls {\em p1} and {\em p2} about two wars war\_x and 
war\_y,
separately. By the result of {\em p1}, it establishes that, to some degree
n people in the country support the war\_x and to some degree m people in the
country  do not support the war\_x, whereas by the result of {\em p2}, 
it establishes
that, to some degree f people in the country support the war\_y and to some
degree g people in the country do not support the war\_y. 
Please note that, 
truth-degree and falsity-degree give a quantitative
description of the {\em supportness} of a poll w.r.t. a war, {\em i.e.}
the supportness is handled as a neutrosophic concept. So, let us consider
\begin{eqnarray*}
\Sigma &=& \{\langle p1:\exists Support.war\_x:\geq 0.6, \leq 0.5 \rangle,
\langle p2:\exists Support.war\_y:\geq 0.8, \leq 0.1 \rangle, \\
 & & war\_x \prec^n War,
war\_y \prec^n War\}
\end{eqnarray*}
where the axioms specify that both war\_x and war\_y are a War. According
to the expansion process, $\Sigma$ will be replaced by
\begin{eqnarray*}
\Sigma^{'} &=& \{\langle p1:\exists Support.war\_x:\geq 0.6, \leq 0.5 \rangle, 
\langle p2:\exists Support.war\_y:\geq 0.8, \leq 0.1 \rangle, \\ 
 & & war\_x :\approx^n  War \sqcap war\_x^*, war\_y :\approx^n War \sqcap 
war\_y^*\},
\end{eqnarray*}
which will be simplified to
\begin{eqnarray*}
\Sigma^{"} &=& \{\langle p1:\exists Support.(War \sqcap war\_x^*):\geq 0.6, 
\leq 0.5 \rangle, \\
 & & \langle p2:\exists Support.(War \sqcap war\_y^*):\geq0.8, 
\leq 0.1 \rangle\}.
\end{eqnarray*}
Now, if we are looking for supportness of polls of War, then from $\Sigma$ we
may infer that 
$\Sigma \models^n \langle p1:\exists Support.War:\geq 0.6, \leq 0.5
 \rangle$ and 
$\Sigma \models^n \langle p2:\exists Support.War:\geq 0.8, \leq 0.1
\rangle$. Furthermore, it is easily verified that 
$\Sigma^{"} \models^n \langle
p1:\exists Support.War:\geq 0.6, \leq 0.5 \rangle$ and $\Sigma^{"} \models^n 
\langle p2:\exists Support.War:\geq 0.8, \leq 0.1 \rangle$ hold as well.
Indeed, for any
neutrosophic assertion $\varphi$, $\Sigma \models^n \varphi$ iff
$\Sigma^{"} \models^n \varphi$ holds. $\hfill \Box$
\end{example}

\section{Decision Algorithms in Neutrosophic ${\cal ALC}$}
Deciding whether $\Sigma \models^n \langle \alpha:\geq n, \leq m \rangle$ or
$\Sigma \models^n \langle \alpha:\leq n, \geq m \rangle$ requires a calculus.
Without loss of generality we will consider purely assertional neutrosophic
KBs only.

We will develop a calculus in the style of the constraint propagation method,
as this method is usually proposed in the context of DLs\cite{18} and fuzzy
DLs\cite{9,11}. We first address the entailment problem, then the subsumption
problem and finally the BTVB problem. Both the subsumption problem and
the BTVB problem will be reduced to the entailment problem.

\subsection{A Decision Procedure for the Entailment Problem}
Consider a new alphabet of ${\cal ALC}$ {\em variables}. An interpretation
is extended to variables by mapping these into elements of the interpretation
domain. An ${\cal ALC}$ {\em object} (denoted by $\omega$) is either an 
individual or a variable.\footnote{In the following, if there is no ambiguity,
${\cal ALC}$ variables and ${\cal ALC}$ objects are called variables and 
objects, respectively.}

A {\em constraint} (denoted by $\alpha$ is an expression of the form
$C(\omega)$ or $R(\omega, \omega^{'})$, where $\omega, \omega^{'}$ are
objects, $C$ is an ${\cal ALC}$ concept and $R$ is a role. A {\em neutrosophic
constraint} (denoted by $\varphi$) is an expression having one of the following
four forms: $\langle \alpha:\geq n, \leq m \rangle, \langle \alpha:\leq n, \geq m \rangle, \langle \alpha:> n, < m \rangle, \langle \alpha:<n, > m \rangle$.
Note that neutrosophic assertions are neutrosophic constraints.

The definitions of satisfiability of a constraint, a neutrosophic constraint, 
a set of constraints, a set of
neutrosophic constraints, atomic constraint and atomic neutrosophic constraint
are obvious.

It is quite easily verified that the neutrosophic entailment problem can be
reduced to the unsatisfiability problem of a set of neutrosophic constraints:
\begin{eqnarray}
\Sigma \models^n \langle \alpha:\geq n, \leq m \rangle &\mbox{iff}& \Sigma \cup \{\langle \alpha:< n, > m \rangle \} \mbox{ not satisfiable} \\
\Sigma \models^n \langle \alpha:\leq n, \geq m \rangle &\mbox{iff}& \Sigma \cup \{\langle \alpha:> n, < m \rangle \} \mbox{ not satisfiable}
\end{eqnarray}
Our calculus, determining whether a finite set $S$ of neutrosophic constraints
is satisfiable or not, is based on a set of constraint propagation rules
transforming a set $S$ of neutrosophic constraints into ``simpler" 
satisfiability preserving sets $S_i$ until either all $S_i$ contain a clash
(indicating that from all the $S_i$ no model of $S$ can be build) or some $S_i$
is completed and clash-free, that is, no rule can be further applied to $S_i$
and $S_i$ contains no clash (indicating that from $S_i$ a model of $S$ can be
build).

A set of neutrosophic constraints $S$ contains a {\em clash} iff it contains
either one of the constraints in Table 1 or $S$ contains a conjugated pair
of neutrosophic constraints. Each entry in Table 2 says us under which 
condition the row-column pair of neutrosophic constraints is a {\em conjugated
pair}.
\begin{table}[htb]
\begin{center}
\begin{tabular}{ll}
$\langle \bot(\omega):\geq n, \leq m \rangle$, & where $n > 0$ or $m < 1$ \\
$\langle \top(\omega):\leq n, \geq m \rangle$, & where $n < 1$ or $m > 0$ \\
$\langle \bot(\omega):> n, < m \rangle$, & $\langle \top(\omega):< n, > m \rangle$ \\
$\langle C(\omega):< 0, > m \rangle$, & 
$\langle C(\omega):> 1, < m \rangle, \langle C(\omega):< n, > 1 \rangle, \langle C(\omega):> n, < 0 \rangle$
\end{tabular}
\end{center}
\caption{Clashes}
\end{table}
\begin{table}[htb]
\begin{center}
\begin{tabular}{|c|||c|c|} 
\hline
& $\langle \alpha:< f, > g \rangle$ & $\langle \alpha:\leq f, \geq g$ \\ \hline
$\langle \alpha:\geq n, \leq m$ & $n \geq f$ or $m \leq g$ & $n > f$ or $m < g$ \\ \hline
$\langle \alpha:> n, < m$ & $n \geq f$ or $m \leq g$ & $n \geq f$ or $m \leq g$ \\ \hline
\end{tabular}
\end{center}
\caption{Conjugated Pairs}
\end{table}
Given a neutrosophic constraint $\varphi$, with $\varphi^c$ we indicate a
conjugate of $\varphi$ (if there exists one). Notice that a conjugate of a 
neutrosophic constraint may be not unique, as there could be infinitely many.
For instance, both $\langle C(a):< 0.6, > 0.3 \rangle$ and $\langle C(a):\leq 0.7, \geq 0.4 \rangle$ are conjugates of $\langle C(a):\geq 0.8, \leq 0.1 \rangle$.

Concerning the rules, for each connective $\sqcap, \sqcup, \neg, \forall, \exists$ there is a rule for each relation $\langle \geq, \leq \rangle, \langle >, < \rangle, \langle \leq, \geq \rangle, \langle <, > \rangle$, {\em i.e.} there are
20 rules. The rules have the form:
\begin{equation}
\Phi \rightarrow \Psi \mbox{ if } \Gamma
\end{equation} 
where $\Phi$ and $\Psi$ are sequences of neutrosophic constraints and $\Gamma$
is a condition. A rule fires only if {\em the condition $\Gamma$ holds, if
the current set $S$ of neutrosophic constraints contains neutrosophic constraints matching the precondition $\Phi$ and the consequence $\Psi$ is not already in
$S$}. After firing, the constraints from $\Psi$ are added to $S$. The rules
are the following:
\begin{eqnarray}
(\neg_{\langle \geq, \leq \rangle}) &  & \langle \neg C(\omega):\geq n, \leq m \rangle \rightarrow \langle C(\omega):\leq m, \geq n \rangle \nonumber \\
(\neg_{\langle >, < \rangle}) &  & \langle \neg C(\omega):> n, < m \rangle \rightarrow \langle C(\omega):< m, > n \rangle \\
(\neg_{\langle \leq, \geq \rangle}) &  & \langle \neg C(\omega):\leq n, \geq m \rangle \rightarrow \langle C(\omega):\geq m, \leq n \rangle \nonumber \\
(\neg_{\langle <, > \rangle}) &  & \langle \neg C(\omega):< n, > m \rangle \rightarrow \langle C(\omega):> m, < n \rangle \nonumber 
\end{eqnarray}
\begin{eqnarray*}
(\sqcap_{\langle \geq, \leq \rangle}) &  & \langle (C \sqcap D)(\omega):\geq n, \leq m \rangle \rightarrow \langle C(\omega):\geq n, \leq m \rangle, \langle D(\omega):\geq n, \leq m \rangle \\
(\sqcap_{\langle >, < \rangle}) &  & \langle (C \sqcap D)(\omega):> n, < m \rangle \rightarrow \langle C(\omega):> n, < m \rangle, \langle D(\omega):> n, < m \rangle \\
(\sqcap_{\langle \leq, \geq \rangle}) &  & \langle (C \sqcap D)(\omega):\leq n, \geq m \rangle \rightarrow \langle C(\omega):\leq n, \geq m \rangle, \langle D(\omega):\geq n, \leq m \rangle | \\
&  & \langle C(\omega):\geq n, \leq m \rangle, \langle D(\omega):\leq n, \geq m \rangle | \\
&  & \langle C(\omega):\leq n, \geq 0 \rangle, \langle C(\omega):\geq 0, \leq m \rangle, \langle D(\omega):\geq n, \leq 1 \rangle, \langle D(\omega):\leq 1, \geq m \rangle | \\
&  & \langle C(\omega):\geq n, \leq 1 \rangle, \langle C(\omega):\leq 1, \geq m \rangle, \langle D(\omega):\geq 0, \leq m \rangle, \langle D(\omega):\leq n, \geq 0 \rangle \\
(\sqcap_{\langle <, > \rangle}) &  & \langle (C \sqcap D)(\omega):< n, > m \rangle \rightarrow \langle C(\omega):< n, > m \rangle, \langle D(\omega):\geq n, \leq m \rangle | \\
&  & \langle C(\omega):\geq n, \leq m \rangle, \langle D(\omega):< n, > m \rangle | \\
&  & \langle C(\omega):< n, > 0 \rangle, \langle C(\omega) \geq 0, \leq m \rangle, \langle D(\omega):\geq n, \leq 1 \rangle, \langle D(\omega):< 1, > m \rangle | \\
&  & \langle C(\omega):\geq n, \leq 1 \rangle, \langle C(\omega):< 1, > m \rangle, \langle D(\omega):< n, > 0 \rangle, \langle D(\omega):\geq 0, \leq m \rangle 
\end{eqnarray*}
\begin{eqnarray*}
(\sqcup_{\langle \geq, \leq \rangle}) &  & \langle (C \sqcup D)(\omega):\geq n, \leq m \rangle \rightarrow \langle C(\omega):\geq n, \leq m \rangle, \langle D(\omega):\leq n, \geq m \rangle | \\
&  & \langle C(\omega):\leq n, \geq m \rangle, \langle D(\omega):\geq n, \leq m \rangle | \\
&  & \langle C(\omega):\geq n, \leq 1 \rangle, \rangle C(\omega):\leq 1, \geq m \rangle, \langle D(\omega):\leq n, \geq 0 \rangle, \langle D(\omega):\geq 0, \leq m \rangle | \\
&  & \langle C(\omega):\geq 0, \leq m \rangle, \langle C(\omega):\leq n, \geq 0 \rangle, \langle D(\omega):\geq n, \leq 1 \rangle, \langle D(\omega):\leq 1, \geq m \rangle \\
(\sqcup_{\langle >, < \rangle}) &  & \langle (C \sqcup D)(\omega):> n, < m \rangle \rightarrow \langle C(\omega):>n, < m \rangle, \langle D(\omega):\leq n, \geq m \rangle | \\
&  & \langle C(\omega):\leq n, \geq m \rangle, \langle D(\omega):> n, < m \rangle | \\
&  & \langle C(\omega):> n, < 1 \rangle, \langle C(\omega):\leq 1, \geq m \rangle, \langle D(\omega):\leq n, \geq 0 \rangle, \langle D(\omega):> 0, < m \rangle | \\
&  & \langle C(\omega):\leq n, \geq 0 \rangle, \langle C(\omega):> 0, < m \rangle, \langle D(\omega):> n, < 1 \rangle, \langle D(\omega):\leq 1, \geq m \rangle \\
(\sqcup_{\langle \leq, \geq}) &  & \langle (C \sqcup D)(\omega):\leq n, \geq m \rangle \rightarrow \langle C(\omega):\leq n, \geq m \rangle, \langle D(\omega):\leq n, \geq m \rangle \\
(\sqcup_{\langle <, >}) &  & \langle (C \sqcup D)(\omega):< n, > m \rangle \rightarrow \langle C(\omega):< n, > m \rangle, \langle D(\omega):< n, > m \rangle 
\end{eqnarray*}
\begin{eqnarray*}
(\forall_{\langle \geq, \leq}) &  & \langle (\forall R.C)(\omega_1):\geq n, \leq m \rangle, \langle R(\omega_1, \omega_2):\geq f, \leq g \rangle \rightarrow \langle C(\omega_2):\geq n, \leq m \rangle \\
&  & \mbox{ if } f > m \mbox{ and } g < n \\
(\forall_{\langle >, <}) &  & \langle (\forall R.C)(\omega_1):>n, < m \rangle, \langle R(\omega_1, \omega_2):\geq f, \leq g \rangle \rightarrow \langle C(\omega_2):>n, < m \rangle \\
&  & \mbox{ if } f \geq m \mbox{ and } g \leq n \\ 
(\exists_{\langle \leq, \geq}) &  & \langle (\exists R.C)(\omega_1):\leq n, \geq m \rangle, \langle R(\omega_1, \omega_2):\geq f, \leq g \rangle \rightarrow \langle C(\omega_2):\leq n, \geq m \rangle \\
&  & \mbox{ if } f > n \mbox{ and } g < m \\
(\exists_{\langle <, >}) &  & \langle (\exists R.C)(\omega_1):< n, > m \rangle,
\langle R(\omega_1, \omega_2):\geq f, \leq g \rangle \rightarrow \langle
C(\omega_2):< n, > m \rangle \\
&  & \mbox{ if } f \geq n \mbox{ and } g \leq m 
\end{eqnarray*}
\begin{eqnarray*}
(\exists_{\geq, \leq}) &  & \langle (\exists R.C)(\omega):\geq n, \leq m \rangle \rightarrow \langle R(\omega, x):\geq n, \leq m \rangle, \langle C(x):\geq n, \leq m \rangle \\
&  & \mbox{ if } x \mbox{ is new variable and there is no } \omega^{'} \mbox{ such that both} \\
&  & \langle R(\omega, \omega^{'}):\geq n, \leq m \rangle \mbox{ and } \langle C(\omega^{'}):\geq n, \leq m \rangle \mbox{ are already in the constraint set}\\
(\exists_{>, <}) &  & \langle (\exists R.C)(\omega):>n, < m \rangle \rightarrow \langle R(\omega, x):>n, < m \rangle, \langle C(x):>n, < m \rangle \\
&  & \mbox{ if } x \mbox{ is new variable and there is no } \omega^{'} \mbox{ such that both} \\
&  & \langle R(\omega, \omega^{'}):>n, < m \rangle \mbox{ and } \langle C(\omega^{'}):> n, < m \rangle \mbox{ are already in the constraint set} \\
(\forall_{\leq, \geq}) &  & \langle (\forall R.C)(\omega):\leq n, \geq m \rangle \rightarrow \langle R(\omega, x):\geq m, \leq n \rangle, \langle C(x):\leq n, \geq m \rangle \\
&  & \mbox{ if } x \mbox{ is new variable and there is no } \omega^{'} \mbox{ such that both } \\
&  & \langle R(\omega, \omega^{'}):\geq m, \leq n \rangle \mbox{ and } \langle C(\omega^{'}):\leq n, \geq m \rangle \mbox{ are already in the constraint set}\\
(\forall_{<, >}) &  & \langle (\forall R.C)(\omega):< n, > m \rangle \rightarrow \langle R(\omega, x):> m, < n \rangle, \langle C(x):< n, > m \rangle \\
&  & \mbox{ if } x \mbox{ is new variable and there is no } \omega^{'} \mbox{ such that both } \\
&  & \langle R(\omega, \omega^{'}):> m, < n  \rangle \mbox{ and } \langle C(\omega^{'}):< n, > m \rangle \mbox{ are already in the constraint set} 
\end{eqnarray*}

A set of neutrosophic constraints $S$ is said to be {\em complete} if no rule
is applicable to it. Any complete set of neutrosophic constraints $S_2$ obtained from a set of neutrosophic constraints $S_1$ by applying the above rules (11)
is called a {\em completion} of $S_1$. Due to the rules $(\sqcup_{\geq, \leq}), (\sqcup_{>, <}), (\sqcap_{\leq, \geq})$ and $(\sqcap_{<, >})$, more than one
completion can be obtained. These rules are called {\em nondeterministic rules}. All other rules are called {\em deterministic rules}.

It is easily verified that the above calculus has the {\em termination property, i.e.} any completion of a finite set of neutrosophic constraints $S$ can be 
obtained after a finite number of rule applications.

\begin{example}
Consider Example 1 and let us prove that $\Sigma^{"} \models^n \langle (\exists Support.War)(p1)$ \\ 
$\geq 0.6, \leq 0.5 \rangle$. We prove the above relation by 
verifying that all completions of 
$S = \Sigma^{"} \cup \{\langle (\exists Support.War)(p1) : < 0.6, > 0.5 \rangle \}$ contain a clash. In fact, we have the 
following sequence.
\begin{table}[htb]
\begin{center}
\begin{tabular}{lll}
$(1)$ & $\langle (\exists Support.(War \sqcap war\_x^*))(p1):\geq 0.6, \leq 0.5 \rangle$ & Hypothesis:$S$ \\
$(2)$ & $\langle (\exists Support.(War \sqcap war\_y^*))(p2):\geq 0.8, \leq 0.1 \rangle$ & \\
$(3)$ & $\langle (\exists Support.War)(p1):< 0.6, > 0.5 \rangle$ & \\ \hline
$(4)$ & $\langle Support(p1, x):\geq 0.6, \leq 0.5 \rangle, \langle (War \sqcap war\_x^*)(x):\geq 0.6, \leq 0.5 \rangle$ & $(\exists_{\geq, \leq}) : (1)$ \\
$(5)$ & $\langle War(x):< 0.6, > 0.5 \rangle$ & $(\exists_{<, >}) : (3), (4)$ \\
$(6)$ & $\langle War(x):\geq 0.6, \leq 0.5 \rangle, \langle war\_x^*(x):\geq 0.6, \leq 0.5 \rangle$ & $(\sqcap_{\geq, \leq}) : (4)$ \\
$(7)$ & clash & $(5), (6)$
\end{tabular}
\end{center}
\end{table}
\end{example}

%\vspace{1in}
$\hfill \Box$
\begin{proposition} 
A finite set of neutrosophic constraints $S$ is satisfiable
iff there exists a clash free completion of $S$. $\hfill \dashv$
\end{proposition}

From a computational complexity point of view, the neutrosophic entailment
problem can be proven to be a PSPACE-complete problem, as is the classical
entailment problem and fuzzy entailment problem.

\begin{proposition}
Let $\Sigma$ be a neutrosophic KB and let $\varphi$ be a neutrosophic assertion. Determining whether $\Sigma \models^n \varphi$ is a PSPACE-complete problem. $\hfill \dashv$
\end{proposition}
\begin{proof}
By the Proposition 1, $\Sigma \models^n \varphi$ iff $\sharp \Sigma \models \sharp \varphi$ and $\star \Sigma \models \star \varphi$. From the PSPACE-completeness of the entailment problem in fuzzy
${\cal ALC}$\cite{11}, PSPACE-completeness of the neutrosophic entailment 
problems follows. $\hfill \Box$
\end{proof}
This result establishes an important property about our neutrosophic DLs.
In effect, it says that no additional computational cost has to be paid
for the major expressive power.

\subsection{A Decision Procedure for the Subsumption Problem}
In this section we address the subsumption problem, {\em i.e.} deciding
whether $C \preceq_{\Sigma_T}^n D$, where $C$ and $D$ are two concepts
and $\Sigma_T$ is a neutrosophic terminology. As we have seen (see Example 1),
$C \preceq_{\Sigma_T}^n D$ can be reduced to the case of an empty terminology
by applying the KB expansion process. So, without loss of generality, we
can limit our attention to the case $C \preceq_{\emptyset}^n D$.

It can easily be shown that

\begin{proposition}
Let $C$ and $D$ be two concepts. It follows that $C \preceq_{\emptyset}^n D$
iff for all $n, m, \langle C(a):\geq n, \leq m \rangle \models^n \langle D(a):\geq n, \leq m \rangle$, where $a$ is a new individual. $\hfill \dashv$
\end{proposition}
\begin{proof}
($\Rightarrow$) Assume that $C \preceq_{\emptyset}^n D$ holds. Suppose to the
contrary that $\exists n, m$ such that $\langle C(a):\geq n, \leq m \rangle \models^n \langle D(a):\geq n, \leq m \rangle$ does not hold. Therefore, 
there is an interpretation ${\cal I}$ and an $n, m$ such that
$|C|^t(a^{\cal I}) \geq n$ and $|D|^t(a^{\cal I}) < n$ or $|C|^f(a^{\cal I}) \leq m$ and $|D|^f(a^{\cal I}) > m$. But, from the hypothesis $n \leq |C|^t(a^{\cal I}) \leq |D|^t(a^{\cal I}) < n$ or $m \geq |C|^f(a^{\cal I}) \geq |D|^f(a^{\cal I}) > m$ follow. Absurd. \\ 
($\Leftarrow$) Assume that for all $n, m, \langle C(a):\geq n, \leq m \rangle \models^n \langle D(a):\geq n, \leq m \rangle$ holds. Suppose to the
contrary that $C \preceq_{\emptyset}^n D$ does not hold. Therefore, there is
an interpretation ${\cal I}$ and $d \in \Delta^{\cal I}$ such that 
$|C|^t(d) > |D|^t(d) \geq 0$ or $|C|^f(d) < |D|^f(d) \leq 1$. Let us extent 
${\cal I}$ to $a$ such that $a^{\cal I} = d$ and consider $\overline{n} = |C|^t(d)$ and $\overline{m} = |C|^f(d)$. Of course, ${\cal I}$ satisfies
$\langle C(a):\geq \overline{n}, \leq \overline{m} \rangle$. Therefore, from
the hypothesis it follows that ${\cal I}$ satisfies 
$\langle D(a):\geq \overline{n}, \leq \overline{m} \rangle$, {\em i.e.} 
$|D|^t(d) \geq \overline{n} = |C|^t(d) > |D|^t(d)$ or 
$|D|^f(d) \leq \overline{m} = |C|^f(d) < |D|^f(d)$. Absurd. $\hfill \Box$
\end{proof}
How can we check whether for all $n, m, \langle C(a):\geq n, \leq m \rangle \models^n \langle D(a):\geq n, \leq m \rangle$ holds? The following proposition shows that

\begin{proposition}
Let $C$ and $D$ be two concepts, $n_1, m_1 \in \{0,0.25,0.5,0.75,1\}$ and let $a$ be an 
individual. It follows that for all $n, m \langle C(a):\geq n, \leq m \rangle \models^n \langle D(a):\geq n, \leq m \rangle$ iff $\langle C(a):\geq n_1, \leq m_1 \rangle \models^n \langle D(a):\geq n_1, \leq m_1 \rangle$ holds. $\hfill \dashv$ 
\end{proposition}
As a consequence, the subsumption problem can be reduced to the entailment problem for which we have a decision algorithm.

\subsection{A Decision Procedure for the BTVB Problem}
We address now the problem of determining $glb(\Sigma, \alpha)$ and 
$lub(\Sigma, \alpha)$. This is important, as computing , {\em e.g.} 
$glb(\Sigma, \alpha)$, is in fact the way to answer a query of type ``to which
degree is $\alpha$ (at least) true and (at most) false, given the (imprecise)
facts in $\Sigma$?".

Without loss of generality, we will assume that all concepts are in NNF ({\em Negation Normal Form}).

\begin{proposition}
Let $\Sigma$ be a set of neutrosophic assertions in NNF and let $\alpha$ be
an assertion. Then $glb(\Sigma, \alpha) \in N^{\Sigma}$ and $lub(\Sigma, \alpha) \in M^{\Sigma}$, where
\[
  N^{\Sigma} = \{\langle n, m \rangle: \langle \alpha:\geq n, \leq m' \rangle \in \Sigma, \langle \alpha:\geq n', \leq m \rangle \in \Sigma\}  
\]
\[
  M^{\Sigma} = \{\langle n, m \rangle: \langle \alpha:\leq n, \geq m' \rangle \in \Sigma, \langle \alpha:\leq n', \geq m \rangle \in \Sigma\}
\]
$\hfill \dashv$
\end{proposition} 

The algorithm computing $glb(\Sigma, \alpha)$ and $lub(\Sigma, \alpha)$ are 
described in Table 3.

\begin{table}[htb]
\begin{center}
\begin{tabular}{|l|}
\hline
{\bf Algorithm} $glb(\Sigma, \alpha)$ \\
Set $Min := \langle 0, 1 \rangle$ and $Max := \langle 1, 0 \rangle$.\\
1. Pick $\langle n, m \rangle \in M^{\Sigma}$ such that {\em first element of Min} $< n <$ {\em first element of Max} and \\
{\em second element of Max} $< m <$ {\em second element of Min}. If there is no such $\langle n, m \rangle$, \\
then set $glb(\Sigma, \alpha) := Min$ and exit. \\ 
2. If $\Sigma \models^n \langle \alpha:\geq n, \leq m \rangle$ then set $Min = \langle n, m \rangle$, else set $Max = \langle n, m \rangle$. Go to Step 1. \\ \\ 
{\bf Algorithm} $lub(\Sigma, \alpha)$ \\
Set $Min := \langle 1, 0 \rangle$ and $Max := \langle 0, 1 \rangle$.\\
1. Pick $\langle n, m \rangle \in N^{\Sigma}$ such that {\em first element of 
Max} $< n <$ {\em first element of Min} and \\
{\em second element of Min} $< m <$ {\em second element of Max}. If there is no
such $\langle n, m \rangle$, \\
then set $lub(\Sigma, \alpha) := Min$ and exit. \\
2. If $\Sigma \models^n \langle \alpha:\leq n, \geq m \rangle$ then set $Min = 
\langle n, m \rangle$, else set $Max = \langle n, m \rangle$. Go to Step 1. \\ 
\hline 
\end{tabular}
\end{center}
\caption{Algorithms $glb(\Sigma, \alpha)$ and $lub(\Sigma, \alpha)$}
\end{table}

\section{Conclusions and Future Work}
In this paper, we have presented a quite general neutrosophic extension of the
fuzzy DL ${\cal ALC}$, a significant and expressive representative of the
various DLs. Our neutrosophic DL enables us to reason in presence of imprecise
(fuzzy, incomplete, and inconsistent) ${\cal ALC}$ concepts, {\em i.e.}
neutrosophic ${\cal ALC}$ concepts. From a semantics point of view, 
neutrosophic concepts are interpreted as neutrosophic sets, {\em i.e.} given
a concept $C$ and an individual $a$, $C(a)$ is interpreted as the truth-value
and falsity-value of the sentence ``$a$ is $C$". From a syntax point of view,
we allow to specify lower and upper bounds of the truth-value and falsity-value
of $C(a)$. Complete algorithms for reasoning in it have been presented, that is,
 we have devised algorithms for solving the entailment problem, the subsumption
problem as well as the best truth-value bound problem.

An important point concerns computational complexity. The complexity result
shows that the additional expressive power has no impact from a computational
complexity point of view.

This work can be used as a basis both for extending existing DL and fuzzy DL
based systems and for further research. In this latter case, there are several
open points. For instance, it is not clear yet how to reason both in case
of neutrosophic specialization of the general form $C \prec^n D$ and in the 
case cycles are allowed in a neutrosophic KB. Another interesting topic for
further research concerns the semantics of neutrosophic connectives. Of course
several other choices for the semantics of the connectives $\sqcap, \sqcup, \neg, \exists, \forall$ can be considered.

\end{document}